\documentclass[letterpaper]{article} 
\usepackage{aaai24}  
\usepackage{times}  
\usepackage{helvet}  
\usepackage{courier}  
\usepackage[hyphens]{url}  
\usepackage{graphicx} 
\usepackage{natbib}  
\usepackage{caption} 
\usepackage{algorithm}
\usepackage[noend]{algpseudocode}
\usepackage{enumitem}
\usepackage{multirow}
\usepackage[table,xcdraw]{xcolor}

\usepackage{mathtools}
\usepackage{amsfonts}
\usepackage{dsfont}
\usepackage[nameinlink]{cleveref}
\usepackage{newfloat}
\usepackage{listings}
\DeclareCaptionStyle{ruled}{labelfont=normalfont,labelsep=colon,strut=off} 
\floatstyle{ruled}
\newfloat{listing}{tb}{lst}{}
\floatname{listing}{Listing}
\title{
Robust Uncertainty Quantification Using Conformalised Monte Carlo Prediction
}
\author{
    Daniel Bethell,
    Simos Gerasimou,
    Radu Calinescu
}
\affiliations{
    Department of Computer Science, University of York\\
    York, United Kingdom\\
    \{daniel.bethell, simos.gerasimou, radu.calinescu\}@york.ac.uk
}

\begin{document}

\maketitle

\begin{abstract}
Deploying deep learning models in safety-critical applications remains a very challenging task, mandating the provision of assurances for the dependable operation of these models. Uncertainty quantification (UQ) methods estimate the model’s confidence per prediction, informing decision-making by considering the effect of randomness and model misspecification. Despite the advances of state-of-the-art UQ methods, they are computationally expensive or produce conservative prediction sets/intervals. We introduce MC-CP, a novel hybrid UQ method that combines a new adaptive Monte Carlo (MC) dropout method with conformal prediction (CP). MC-CP adaptively modulates the traditional MC dropout at runtime to save memory and computation resources, enabling predictions to be consumed by CP, yielding robust prediction sets/intervals. Throughout comprehensive experiments, we show that MC-CP delivers significant improvements over comparable UQ methods, like MC dropout, RAPS and CQR, both in classification and regression benchmarks. MC-CP can be easily added to existing models, making its deployment simple. The MC-CP code and replication package is available at \url{https://github.com/team-daniel/MC-CP}.
\end{abstract}

\section*{Introduction}
Advances in Deep Learning (DL) enable its employment in diverse and challenging tasks, including speech recognition~\cite{aaai-speech-recog} and image annotation~\cite{image-anno-jmlr}. Despite its numerous potential applications, using DL in safety-critical applications (e.g., medical imaging/diagnosis) mandates ensuring its dependable and robust operation~\cite{chall-ml,gerasimou2020importance}. Uncertainty quantification (UQ) is crucial in assessing the DL model's confidence for input-prediction pairs and establishing the potential impact of noisy, sparse, or low-quality input and misspecification in DL models~\cite{bayesian-segnet}. Ultimately, UQ enables understanding situations where the model is particularly uncertain, instrumenting uncertainty-aware decision-making~\cite{calinescu2018efficient}. 

DL-focused methods for UQ aim at assessing model and data uncertainty of DL models~\cite{abdar2021review}. In particular, Monte Carlo (MC) dropout~\cite{mcdropout} elegantly quantifies uncertainty within DL models by outputting the standard deviation of predictions from an ensemble of networks using dropout layers. Running, however, numerous forward passes is computationally expensive. 
Similarly, Bayesian Neural Networks (BNNs)~\cite{MacKay-BNN} constitute a more natural UQ method that can estimate both epistemic and aleatoric uncertainty. However, BNNs are computationally-intensive both during training and inference and require substantial fine-tuning. Finally, conformal prediction (CP)~\cite{cpbook} produces prediction sets/intervals instead of singletons. The larger the set/interval, the more unsure the model is about its prediction, with a singleton prediction/narrow interval typically signifying large confidence. Despite their merits, CP methods are over-conservative, producing larger sets/intervals than necessary~\cite{cp-overestimate}.

Driven by these advances, we introduce \textbf{M}onte \textbf{C}arlo-\textbf{C}onformal \textbf{P}rediction (MC-CP), a novel hybrid method that comprises adaptive MC dropout and conformal predictive techniques, inheriting both the statistical efficiency of the former and the distribution-free coverage guarantee of conformal prediction. MC-CP dynamically adapts the conventional MC dropout with a convergence assessment, saving memory and computational resources during inference where possible. The predictions are then consumed by advanced CP techniques to synthesize robust prediction sets/intervals. Our experimental evaluation shows that the hybrid MC-CP approach overestimates less than regular CP methods. Despite its simplicity, it outperforms state-of-the-art CP- and MC-based methods, e.g.,  traditional MC dropout, RAPS~\cite{raps} and CQR~\cite{cqr}, both in classification and regression benchmarks. While RAPS and CQR quantify uncertainty by increasing the prediction set/interval size, MC-CP does this and also outputs an exact quantification in the form of variance in the prediction distribution. Our MC-CP method is designed to be implemented at inference time, in contrast to evidential deep learning and Bayesian neural networks. Whilst these methods provide salient and informative UQ estimations, MC-CP is realised post-training.

Our contributions are:

\begin{itemize}[noitemsep, nolistsep]
    \item An adaptive MC dropout method that can save computational resources compared to the original method;
    \item The hybrid MC-CP method that addresses major issues common with CP methods, yielding significant improvements across several metrics and datasets.
    \item A comprehensive empirical MC-CP evaluation against state-of-the-art UQ methods (MC Dropout, RAPS, CQR) on various benchmarks, including CIFAR-10, CIFAR-100, MNIST, Fashion-MNIST, and Tiny ImageNet.
\end{itemize}

\noindent
Paper Structure:
Sections~\ref{sec:relatedWork} and~\ref{sec:preliminaries} discuss related UQ work and background material.
Sections~\ref{sec:approach} and~\ref{sec:results} present MC-CP and its empirical evaluation.
Section~\ref{sec:conclusion} concludes the paper.

\section*{Related Work}
\label{sec:relatedWork}
Uncertainty Quantification (UQ) in DL indicates how uncertain a model is about its predictions. The most common uncertainty types are aleatoric and epistemic. The former surrounds the irreducible uncertainty within data (e.g., random noise). The latter is the model's lack of knowledge or poor training which can be reduced with more data or better training. MC-CP focuses on quantifying epistemic uncertainty.

Deep ensembles is a straightforward method to quantify uncertainty in DL~\cite{deep-ensembles}. The method involves training an ensemble of networks with the same or similar architecture, initialised with different weights. After training, the ensemble predicts on the same input data using the mean of their predictions as the final prediction and the variance as the uncertainty.

Monte Carlo (MC) dropout~\cite{mcdropout} is a simple and effective method to compute epistemic uncertainty in DL models by exploiting dropout~\cite{dropout}, a regularization technique that randomly drops units of the neural network to prevent reliance on certain weights. Although dropout is typically used during training, MC dropout keeps this feature active during inference and
performs several forward passes to devise a prediction distribution. The final prediction is the mean of the distribution, and the variance signifies the uncertainty. Gaussian dropout~\cite{gaussian-dropout} complements regular dropout by adding noise using a Gaussian distribution instead of setting the unit's value to zero.

Bayesian Neural Networks (BNNs)~\cite{bayesian-segnet} realise UQ directly in the model's architecture. While in traditional DL networks, weights are a singleton variable, in BNNs, weights are represented as a distribution. Although BNNs produce probabilistic predictions that naturally capture uncertainty, they are computationally intensive and require substantially more training than standard networks, resorting to approximate Bayesian computation techniques like variational inference.

Conformal prediction (CP)~\cite{cpbook} is a framework that uses validity to quantify a model's prediction confidence. Validity encodes that, on average, a model's predictions will be correct within a guaranteed confidence level (e.g., 90\% of the time). The method then alters the prediction from a singleton/point to a set/interval that indicates the confidence level of the model. The larger the set/interval, the more uncertain the model is, and vice versa. CP involves splitting the test data into two sets: a calibration and a test set. The calibration set is used to estimate the thresholds needed to achieve the desired confidence levels. CP has been applied to a diverse set of applications (e.g., image classification~\cite{raps}, regression~\cite{cqr}, object detection~\cite{cp-obj-det}).

An orthogonal method is test time augmentation~\cite{tta-med, tta-med2} which alters the data at inference time instead of the model or predictions. Given an input, the method creates multiple augmented inputs using various augmentation techniques. The DL model then makes predictions for the augmented inputs; their distribution and variance represent the model's uncertainty. 
Data augmentation using generative AI has also been proposed to enhance the inference capabilities of DL models~\cite{missaoui2023semantic}.

\section*{Preliminaries}
\label{sec:preliminaries}
Given a level of coverage $\alpha \in (0,1)$ signifying a probability guarantee that the true label/point is in the prediction set/interval $(1 - \alpha)$\%,  Conformal prediction (CP) constructs a prediction set/interval instead of a singleton/point. To achieve this, CP splits the test dataset into a calibration set $c$ and a validation set $v$. Next, conformal scores $s(f(x_i), y_i) \in \mathbb{R}$ are calculated for each $(x_i, y_i) \in c$. This score is high when the model $f(.)$ produces a low softmax output for the true class, i.e., when the model is very wrong. A quantile threshold $\hat{q} = Q(\frac{[(n+1)(1-\alpha)]}{n})$ is calculated, using the desired coverage $\alpha$, the calibration set $c$, and the size of the calibration set $n$, which is used to form prediction sets $C(x_j) = \{y : f(x_j) \leq 1-\hat{q}\}$ for each new input $x_j$ (e.g., from the validation set $v$). For quantile regression, prediction intervals are formed by $C(x_j) = [{t}_{\alpha/2}(x_j)-\hat{q},{t}_{1-\alpha/2}(x_j)+\hat{q}]$ where ${t}$ are the $\alpha$-informed quantiles produced by the trained model.

Coverage is a key metric for assessing CP, measuring how often the predicted set/interval contains the ground truth. Coverage is expected to reflect the desired coverage property $1 -\alpha$. Given model $f$, coverage is calculated by:
\begin{equation}
Coverage(f_{\alpha}) = \frac{1}{n}\sum^{n}_{i=1}\mathds{1}\{y_{i}\in f_{\alpha}(x_{i})\}
\label{eq:coverage}
\end{equation}

where $\alpha$ is the user-defined coverage, $n$ is the size of the validation set, $y_i$ is the true label/value, and $f_{\alpha}(x)$ is the prediction interval/set made by the model for input $x_i$. This equation reflects the percentage of true labels/values captured by the respective prediction sets/intervals.

Efficiency is another important CP metric. While including all possible classes in a prediction set would, by default, yield a perfect accuracy score, it is impractical. Thus, a DL model that achieves the desired coverage efficiently is preferred. Efficiency is calculated as the average expected size of the set/interval, given by:
\begin{equation}
Size(f_{\alpha}) = \frac{1}{n}\sum^{n}_{i=1}|f_{\alpha}(x_{i})|
\label{eq:efficiency}
\end{equation}

\begin{figure}[t]
    \centerline{\includegraphics[width=0.95\linewidth]{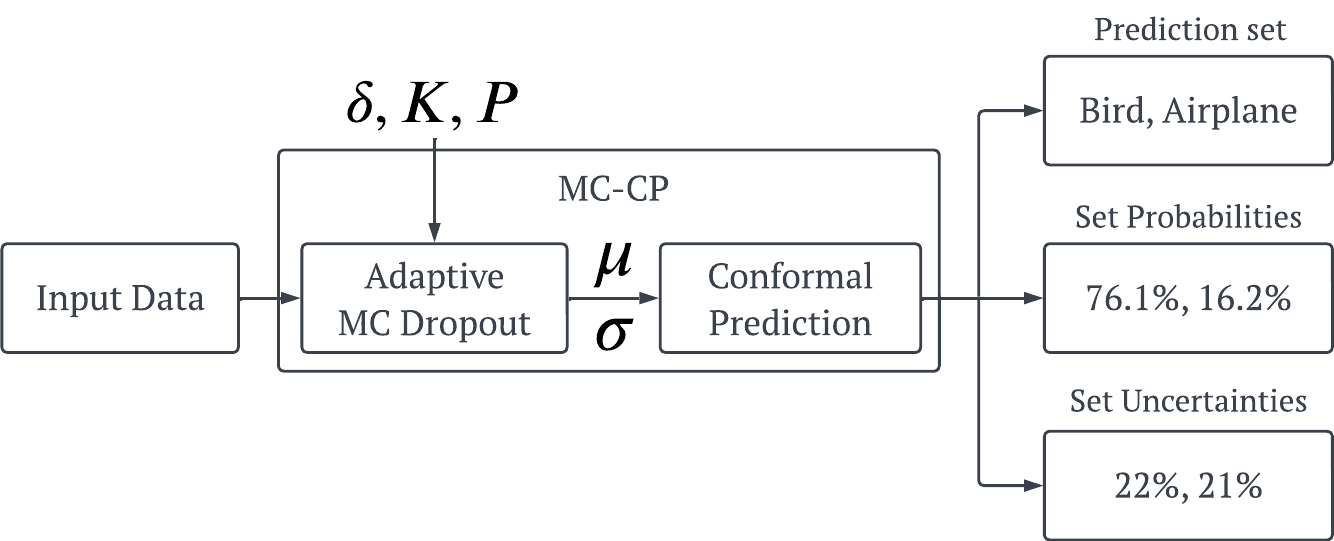}}
    \caption{High-level overview of our MC-CP method for image classification.}
    \label{fig:mccp-overview}
\end{figure}

\section*{MC-CP}
\label{sec:approach}
Our \textbf{M}onte-\textbf{C}arlo \textbf{C}onformal \textbf{P}rediction (MC-CP) method for UQ incorporates adaptive MC dropout and conformal prediction, leveraging their low computational cost and finite sample distribution-free coverage guarantees, respectively. 
Fig.~\ref{fig:mccp-overview} shows a high-level overview of our MC-CP method for image classification.
We discuss next adaptive MC dropout, followed by an exposition of MC-CP for classification and regression. 
This novel combination of adaptive MC dropout and CP, albeit straightforward, results in a hybrid MC-CP method that yields significant improvements compared to state-of-the-art UQ techniques (Section~\ref{sec:results}).

\subsection*{Adaptive Monte Carlo Dropout}
The competitive predictive performance of MC dropout largely depends on the execution of multiple stochastic forward passes of each input through the DL model at inference time.
The number of forward passes $K$ the model should perform is defined a priori and is fixed. 
Since, for any new input, the dropout layers of the DL model are kept on during inference, the ensemble of these $K$ forward passes produces a distribution of predictions.
This distribution enables quantifying uncertainty by computing metrics such as the expected (average) value, standard deviation and entropy.

The motivation underpinning adaptive MC dropout originates from the observation that each forward pass corresponds to a particular DL model instantiation that adds unique variance to the prediction distribution. 
Some of these DL model instantiations, informed by MC dropout forward passes, can produce similar or even the exact same prediction. 
Hence, although the prediction variance might be large initially, as the number of forward passes increases, the variance value becomes smaller, indicating that the inference process has converged. 
If the current number of forward passes is substantially less than the maximum number of forward passes $K$ when this event occurs, the remaining forward passes incur only additional overheads but add little to no value. 
Adaptive MC Dropout leverages this observation to reduce the number of wasted forward passes once convergence is diagnosed, thus yielding significant computational savings without impacting the prediction effectiveness.

\begin{algorithm}[t]
\caption{Adaptive Monte Carlo Dropout}
\label{alg:DMCDropout}
\textbf{Input}: Model $f$, Input $x$, Maximum forward passes $K$, Threshold $\delta$ and Patience $P$\\
\textbf{Output}: Mean prediction $\mu$, and Variance $\sigma$
\begin{algorithmic}[1] 
\State Count $\leftarrow$ 0
\State Predictions $\leftarrow$ []
\While{(Count $<$ $P$ $\&$ size(Predictions) $<$ $K$)}
    \State $y \leftarrow f(x)$
    \State Predictions $\leftarrow$ Predictions $\cup\;y$
    \State $\sigma \leftarrow Var(\text{Predictions})$
    \If {(Predictions $>$ 1)}
        \State $\text{diff} \leftarrow |\sigma_{i-1} - \sigma|$\Comment{list of differences}
        \If{($\forall z \in \text{diff}.z \leq \delta$)}
            \State Count $\leftarrow$ Count + 1
        \Else
            \State Count $\leftarrow$ 0
        \EndIf
    \EndIf
    \State $\sigma_{i-1} \leftarrow \sigma$
\EndWhile
\State \textbf{return} $\overline{\textrm{Predictions}},\; \sigma$
\end{algorithmic}
\end{algorithm}

Algorithm~\ref{alg:DMCDropout} shows our adaptive MC dropout method. 
Given a new input $x$, the method performs up to $K$ forward passes over model $f$ to produce the predictive posterior mean as the final prediction and the variance of the predictive posterior as the prediction uncertainty. 
Unlike conventional MC dropout, our algorithm uses the hyperparameters threshold $\delta$ and patience $P$ to detect the convergence and terminate early. 
The threshold parameter $\delta$ denotes the maximum difference in variance required to trigger that the class/quantile prediction has likely converged. 
Patience $P$ signifies the number of consecutive forward passes where all classes/quantiles are below $\delta$ to stop the execution early. 
The criterion of performing $P$ successive forward passes that meet the threshold $\delta$ is important in determining convergence and mitigating the potential effect of randomness.

Adaptive MC dropout works as follows. While the current forward pass counter is less than $K$ and the current patience counter is less than $P$ (line 3), the model predicts the input data with dropout layers switched on (line 4). 
The prediction is added to a list, and the variance of that list is estimated (lines 5-6). 
From the second forward pass onward, the difference between the current variance $\sigma$ and the last estimated variance $\sigma_{i-1}$ is calculated (line 8). 
If the difference for all classes/quantiles is below the threshold $\delta$, then the current patience counter is increased (lines 9-10); otherwise, it is reset (line 12). 
Once all classes/quantiles converge below $\delta$  after $P$ consecutive forward passes, the predictive posterior mean and variance are outputted as the predictions and their measured uncertainty, respectively (line 14). 

The user-defined parameters threshold $\delta \in (0, 1)$ and patience $P \in \mathbb{Z_+}$ enable controlling the sensitivity of the adaptive MC dropout to changes in prediction variance. 
When $\delta$ approaches 1, our method becomes less sensitive, allowing to stop earlier. 
In contrast, the closer $\delta$ is to 0, the more sensitive it becomes, requiring the execution of more forward passes until convergence is diagnosed. 
It can be easily seen that selecting a small $\delta$ and large patience $P$ values enables instrumenting the conventional MC dropout method. 
We demonstrate this remark later in~\Cref{tab:various-delta,tab:reg-sens-analysis}.

We also provide a sketch of the proof for the adaptive MC dropout method. The MC Dropout process is a Bernoulli process; each MC Dropout forward pass is independent of the others, and the model parameters are fixed during our adaptive MC Dropout approach. According to the Law of Large Numbers, as the number of $\mathrm{Predictions}$ from line~5 of Algorithm~1 increases, the sample variance $\sigma$ from line~6 will converge to the true variance $\sigma_{true}$ of the MC Dropout output population, and there exists a number of forward passes $N=\#\mathrm{Predictions}$ such that for all $ i \geq N$, $|\sigma - \sigma_{true}| < \delta/2$. We show that the while loop from lines 3--13 terminates after fewer than $K$ iterations if $N<K-P$. To that end, we note that, since the $\sigma$ value computed in iterations $N$, $N+1$, \ldots, $N+P$ of the while loop is within $\delta/2$ of $\sigma_{true}$, in each of these successive iterations  $\mathrm{diff}=|\sigma_{i-1}-\sigma|<\delta$ in line~8, and therefore $\mathrm{Count}$ is incremented in line~10, reaching the value $P$ and ending the while loop before $K$ iterations.

\begin{algorithm}[t]
\caption{MC-CP for image classification}
\label{alg:mccpImageClassification}
\textbf{Input}: Model $f$, Test set, Maximum forward passes $K$, Threshold $\delta$, and Patience $P$ \\
\textbf{Output}: Prediction set, and variance set\\
\textit{\textbf{Conformal Calibration}}
\begin{algorithmic}[1] 
\State \textbf{Split test set:} split the test set in calibration $c$ and validation $v$.
\State \textbf{Calibrate:} perform Platt scaling on the model using $c$.
\State \textbf{Calculate conformal score:} For each image in the training set, define $E_{j} = \sum^{k'}_{i=1}(\hat{\pi}_{(i)}(x_{j})+\lambda1[i>k_{reg}])$ where $k'$ is the model's ranking of the true class $y_{j}$ and $\hat{\pi}_{(i)}(x_{j})$ is the $i^{th}$ largest score for the $j^{th}$ image.
\State \textbf{Find the threshold:} assign $\hat{T}_{ccal}$ to the $1 - \alpha$ quantile of the $E_{j}$.
\end{algorithmic}
\textit{\textbf{Conformal Prediction}}
\begin{algorithmic}[1] 
\State \textbf{Mean softmax:} retrieve softmax and variance from Adaptive Monte Carlo Dropout($f, v, K, \delta, P)$.
\State \textbf{Prediction set:} output the $k^{*}$ highest-score classes, where $E^{k^{*}}_{i=1} = \sum^{k'}_{i=1}(\hat{\pi}_{(i)}(x_{n+1})+\lambda1[j>k_{reg}]) \geq \hat{T}_{ccal}$.
\end{algorithmic}
\end{algorithm}

\subsection*{MC-CP for Image Classification}
For image classification, we combine our Adaptive Monte Carlo dropout method with conformal prediction to form MC-CP, shown in Algorithm~\ref{alg:mccpImageClassification}. MC-CP is split into two steps, conformal calibration and prediction. First, a test dataset is split into calibration and validation sets. Platt scaling is then performed on the pre-trained model using the calibration dataset. Next, we calculate the conformal scores for each input image in the training set, which can then be used to calculate the quantile threshold $\hat{q}$.

During the prediction stage of MC-CP, we invoke the adaptive MC dropout method, with the selected hyperparameters, for each new input image. 
This invocation returns the mean prediction and variance of the possible classes of the image. The final prediction set can then be determined by calculating the cumulative softmax output for all classes and then including the classes from most to least likely that do not exceed the quantile threshold. In Section~\ref{sec:results}, we show how MC-CP outperforms other state-of-the-art conformal prediction techniques, with modest computational overheads.

\subsection*{MC-CP for Regression}
We also develop an extension of MC-CP for deep quantile regression, shown in Algorithm~\ref{alg:mccpRegression}. This is also split up into calibration and prediction steps. To calculate the conformal scores, the magnitude of error for the desired quantiles is estimated. Next, the threshold can be calculated using the calibration dataset.

For the prediction stage of MC-CP for deep quantile regression, once again, the adaptive MC Dropout method is called, with the desired hyperparameters, for each data point in the validation dataset. Finally, a prediction interval is calculated for both quantiles on an unseen data point in the validation set using the calculated threshold.
In Section~\ref{sec:results}, we show how MC-CP outperforms regular deep quantile regression and the CQR method.

\begin{algorithm}[t]
\caption{MC-CP for deep quantile regression}
\label{alg:mccpRegression}
\textbf{Input}: Model $f$, Test set, Maximum ensemble $K$, Threshold $\delta$, and Patience $P$\\
\textbf{Output}: Prediction interval, and variance\\
\textit{\textbf{Conformal Calibration}}
\begin{algorithmic}[1] 
\State \textbf{Split test set:} split the test set in calibration $c$ and validation $v$.
\State \textbf{Calculate conformal score:} for each data point in $c$, define $E_{i} \coloneqq max\{\hat{q}_{\alpha_{lo}}(x_{i})-y_{i}, y_{i} - \hat{q}_{\alpha_{hi}}(x_{i})\}$.
\State \textbf{Find the threshold:} compute $Q_{q-\alpha}(E, c)$, the $(1-\alpha)(1+1/|c|)$-th empirical quantile of $\{E_{i}:i\in c\}$.
\end{algorithmic}
\textit{\textbf{Conformal Prediction}}
\begin{algorithmic}[1] 
\State \textbf{Mean softmax:} retrieve softmax and variance from Adaptive Monte Carlo Dropout($f, v, K, \delta, P)$.
\State \textbf{Prediction Interval:} output the prediction interval $C(v)= [\hat{q}_{\alpha_{lo}}(v) - Q_{1-\alpha}(E, c), \hat{q}_{\alpha_{hi}}(v) + Q_{1-\alpha}(E, c)]$ for unseen validation data $v$.
\end{algorithmic}
\end{algorithm}

\section*{Evaluation}
\label{sec:results}

\subsection*{Experimental Setup}
\noindent
\textbf{Benchmarks.} 
For classification, we evaluate MC-CP on five image datasets: CIFAR-10 and CIFAR-100~\cite{CIFAR}, MNIST~\cite{MNIST}, Fashion-MNIST~\cite{fashion-mnist}, and Tiny ImageNet~\cite{tinyImageNet}. CIFAR-10 and CIFAR-100 contain $60,000$ 32x32 colour images with $10$ and $100$ classes respectively. MNIST and Fashion-MNIST contain $60,000$ 28x28 grey-scale images with $10$ classes each. Tiny ImageNet is a small version of the well-known ImageNet dataset containing $100,000$ 64x64 colour images with $200$ classes.

For regression, we use the following five benchmarks: Boston Housing~\cite{boston-housing}, Abalone~\cite{abalone}, Blog Feedback~\cite{blog-feedback}, Concrete Compressive Strength~\cite{concrete}, and Physicochemical Properties of Protein Tertiary Structure dataset~\cite{protein}. The Boston Housing dataset contains 506 data points with 14 attributes, the Abalone dataset has 4180 data points with 9 attributes, the Blog Feedback dataset contains 60,021 data points with 281 attributes, the Concrete dataset contains 1,030 data points with 9 attributes, and the Physicochemical Properties of Protein Tertiary Structure dataset contains 45,730 data points with 9 attributes.

\begin{table}[t]
\renewcommand{\arraystretch}{1.25}
\centering
\resizebox{\columnwidth}{!}{%
\begin{tabular}{|l|l|l|l|}
\hline
\multicolumn{1}{|c|}{\textbf{Dataset}} & \multicolumn{1}{c|}{\textbf{Tech.}} & \textbf{Test Error} & \textbf{Pred Sizes} \\ \hline
\rowcolor[HTML]{F3F3F3} 
\cellcolor[HTML]{F3F3F3} & Baseline & 38.05 $\pm$ 0.36 & 1.00 $\pm$ 0.00 \\ \cline{2-4} 
\rowcolor[HTML]{F3F3F3} 
\cellcolor[HTML]{F3F3F3} & MC & 35.54 $\pm$ 0.36 & 1.00 $\pm$ 0.00 \\ \cline{2-4} 
\rowcolor[HTML]{F3F3F3} 
\cellcolor[HTML]{F3F3F3} & Naive & 5.07 $\pm$ 0.28 & 3.53 $\pm$ 1.63 \\ \cline{2-4} 
\rowcolor[HTML]{F3F3F3} 
\cellcolor[HTML]{F3F3F3} & RAPS & 3.29 $\pm$ 0.30 & 4.35 $\pm$ 1.86 \\ \cline{2-4} 
\rowcolor[HTML]{F3F3F3} 
\multirow{-5}{*}{\cellcolor[HTML]{F3F3F3}\textbf{CIFAR-10}} & MC-CP & 1.47 $\pm$ 0.15 & 4.11 $\pm$ 1.81 \\ \hline
 & Baseline & 72.14 $\pm$ 0.87 & 1.00 $\pm$ 0.00 \\ \cline{2-4} 
 & MC & 69.46 $\pm$ 0.60 & 1.00 $\pm$ 0.00 \\ \cline{2-4} 
 & Naive & 4.92 $\pm$ 0.42 & 39.25 $\pm$ 11.43 \\ \cline{2-4} 
 & RAPS & 4.83 $\pm$ 0.25 & 41.65 $\pm$ 4.01 \\ \cline{2-4} 
\multirow{-5}{*}{\textbf{CIFAR-100}} & MC-CP & 3.54 $\pm$ 0.20 & 39.26 $\pm$ 3.95 \\ \hline
\rowcolor[HTML]{F3F3F3} 
\cellcolor[HTML]{F3F3F3} & Baseline & 1.10 $\pm$ 0.06 & 1.00 $\pm$ 0.00 \\ \cline{2-4} 
\rowcolor[HTML]{F3F3F3} 
\cellcolor[HTML]{F3F3F3} & MC & 1.11 $\pm$ 0.04 & 1.00 $\pm$ 0.00 \\ \cline{2-4} 
\rowcolor[HTML]{F3F3F3} 
\cellcolor[HTML]{F3F3F3} & Naive & 5.01 $\pm$ 0.58 & 0.95 $\pm$ 0.22 \\ \cline{2-4} 
\rowcolor[HTML]{F3F3F3} 
\cellcolor[HTML]{F3F3F3} & RAPS & 1.10 $\pm$ 0.04 & 1.08 $\pm$ 0.35 \\ \cline{2-4} 
\rowcolor[HTML]{F3F3F3} 
\multirow{-5}{*}{\cellcolor[HTML]{F3F3F3}\textbf{MNIST}} & MC-CP & 0.32 $\pm$ 0.01 & 1.06 $\pm$ 0.33 \\ \hline
 & Baseline & 12.01 $\pm$ 0.26 & 1.00 $\pm$ 0.00 \\ \cline{2-4} 
 & MC & 12.08 $\pm$ 0.23 & 1.00 $\pm$ 0.00 \\ \cline{2-4} 
 & Naive & 4.93 $\pm$ 0.40 & 1.20 $\pm$ 0.42 \\ \cline{2-4} 
 & RAPS & 1.12 $\pm$ 0.07 & 1.80 $\pm$ 0.99 \\ \cline{2-4} 
\multirow{-5}{*}{\textbf{Fashion-MNIST}} & MC-CP & 0.82 $\pm$ 0.06 & 1.76 $\pm$ 1.00 \\ \hline
\rowcolor[HTML]{F3F3F3} 
\cellcolor[HTML]{F3F3F3} & Baseline & 81.07 $\pm$ 1.05 & 1.00 $\pm$ 0.00 \\ \cline{2-4} 
\rowcolor[HTML]{F3F3F3} 
\cellcolor[HTML]{F3F3F3} & MC & 78.60 $\pm$ 2.37 & 1.00 $\pm$ 0.00 \\ \cline{2-4} 
\rowcolor[HTML]{F3F3F3} 
\cellcolor[HTML]{F3F3F3} & Naive & 4.85 $\pm$ 0.34 & 97.53 $\pm$ 29.64 \\ \cline{2-4} 
\rowcolor[HTML]{F3F3F3} 
\cellcolor[HTML]{F3F3F3} & RAPS & 4.57 $\pm$ 0.09 & 107.78 $\pm$ 2.06 \\ \cline{2-4} 
\rowcolor[HTML]{F3F3F3} 
\multirow{-5}{*}{\cellcolor[HTML]{F3F3F3}\textbf{Tiny ImageNet}} & MC-CP & 3.99 $\pm$ 0.41 & 97.17 $\pm$ 3.67 \\ \hline

\end{tabular}%
}
\caption{Test errors (\%) and prediction sizes per UQ method on the five classification benchmarks ($\delta$=5e-4, $P$=10).}
\label{tab:test-errors}
\end{table}

\vspace{1mm}\noindent
\textbf{UQ Methods Configuration.} 
In our classification experiments, all methods use a basic convolution neural network (CNN) architecture comprising two hidden layers, two pooling layers, and two dropout layers with a frequency of 50\%. All models are trained on a batch size of 128 for 10 epochs. The categorical cross entropy loss function and stochastic gradient descent optimiser with a learning rate and momentum of $0.1$ and $0.9$, respectively. Each experiment is repeated five times to account for stochasticity. For CP methods, the calibration set size is 25\% of the test set and $\alpha = 0.05$. We do not consider Deep Ensembles or Bayesian Neural Networks within our experiments. These techniques require heavy fine-tuning between datasets, disallowing us to establish a clear baseline. It would not be evident if performance differences would be due to hyperparameter tuning or the method itself. 
To enable a fair comparison, we use the same network architecture and hyperparameters for all classification-based UQ methods, i.e.,  a standard CNN, a CNN with MC dropout, Naive CP, RAPS, and MC-CP (instrumented with RAPS).

In our regression experiments, all methods use a deep quantile regression model comprising two hidden layers and two dropout layers with a frequency of 25\%. The learning rate of the Adam optimiser is $0.001$, and a custom multi-quantile loss function is used with the quantiles $0.05$ and $0.95$. Each model is trained for 100 epochs on a batch size of 32, with experiments repeated five times to consider stochasticity. For CP methods, the calibration set size is 2\% of the test set and $\alpha = 0.1$.
As before, the same DL model is used for comparing the following regression-based UQ methods: a deep quantile regressor, a deep quantile regressor with MC dropout, CQR, and MC-CP instrumented with CQR.

\begin{figure}[t]
    \centerline{\includegraphics[width=\linewidth]{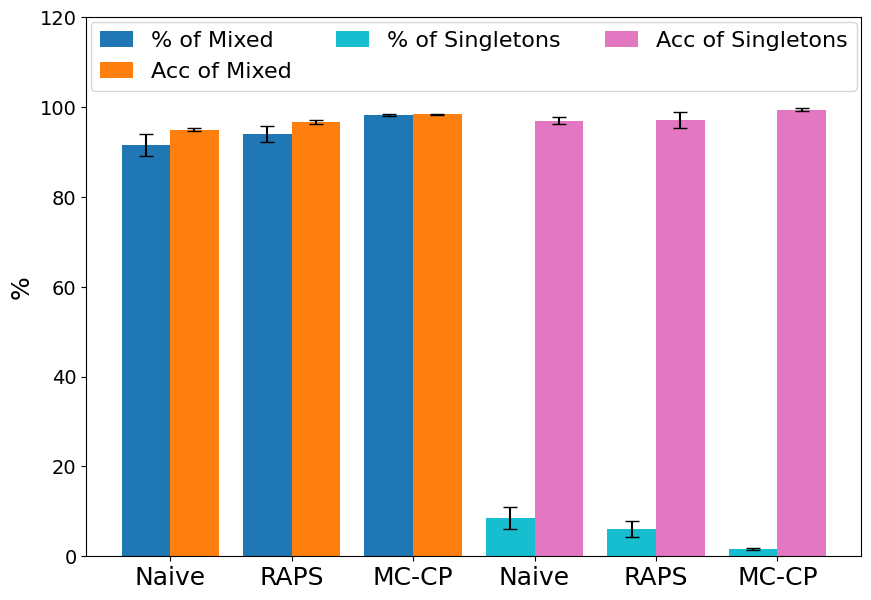}}
    \caption{Percentage and accuracy of singleton and mixed predictions for Naive CP, RAPS, MC-CP on CIFAR-10.}
    \label{fig:singleton-metrics}
\end{figure}

\subsection*{Image Classification Results}
\textbf{Classification Accuracy.} The accuracy results of five different methods against each of the datasets are shown in Table~\ref{tab:test-errors}. The methods tested against MC-CP were a baseline CNN, the same CNN with MC dropout applied, Naive conformal prediction~\cite{conformal-prediction-gentle-intro}, and RAPS. Results show that not only does MC-CP have increased accuracy in comparison to baseline and state-of-the-art conformal prediction methods, but it also does so with less deviation between runs. In particular, we emphasise that our method consistently increases accuracy and yields a lower standard deviation on difficult datasets such as CIFAR-10, CIFAR-100 and Tiny ImageNet. 
Further, and as expected, conformal prediction methods can drastically improve accuracy compared to baseline methods, such as regular CNN and MC dropout. However, MC-CP improves accuracy substantially with less deviation between runs, highlighting its consistency with Naive CP and RAPS.

\vspace{1mm}\noindent
\textbf{Singleton and Mixed Predictions.} Next, we compare the percentage and accuracy of singleton and non-singleton (mixed) predictions for all three conformal prediction methods on CIFAR-10 (Figure~\ref{fig:singleton-metrics}). Naive CP is more likely to predict singleton values, whereas our method is least likely.
When a model is not confident about its prediction, CP-based methods should desirably increase the prediction set size to account for this uncertainty and, hopefully, include the correct class in the larger prediction set. 
The comparison of singleton and non-singleton results in Figure~\ref{fig:singleton-metrics} provides evidence that our method correctly increases the set size to improve accuracy. In fact, for both singleton and non-singleton set sizes, our method performs with the highest accuracy, also exhibiting a consistent behaviour, as indicated by the 
low amount of variance between runs. 

An argument can be made that making the set size large enough could cover nearly all the classes, and this behaviour could reflect a higher accuracy. Comparing these results with the mean set sizes in Table~\ref{tab:test-errors}, we can see that all methods only cover a portion of the classes in their mean set sizes.

\begin{figure}[t]
    \centerline{\includegraphics[width=0.95\linewidth]{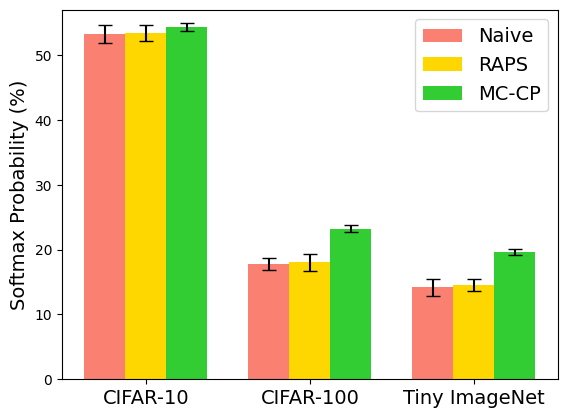}}
    \caption{Mean confidence of top predictions for Naive CP, RAPS, MC-CP on CIFAR-10, CIFAR-100, Tiny ImageNet.}
    \label{fig:mean-cred}
\end{figure}

\vspace{1mm}\noindent
\textbf{Confidence of Predictions.}
We evaluated whether MC-CP could result in a more confident model than traditional conformal prediction methods, thus providing improved accuracy. 
Figure~\ref{fig:mean-cred} shows the mean highest softmax output for every CP method for CIFAR-10, CIFAR-100, and Tiny ImageNet. 
Compared to Naive CP and RAPS, our method shows an increase in confidence across all benchmarks. 
Looking closely at larger-scale datasets, such as CIFAR-100 and Tiny ImageNet, MC-CP is substantially more confident in its predictions. We also observe, in Figure~\ref{fig:mean-cred}, that MC-CP consistently has a smaller standard deviation between runs than the other methods.

\vspace{1mm}\noindent
\textbf{Prediction Sets Size.} 
We have already shown how the accuracy of each method has been tested at scale using CIFAR-100. However, this only reflects a portion of the performance of each method at scale and doesn't highlight any of its weaknesses. The `Prediction Sizes' column in Table~\ref{tab:test-errors} shows the mean set size and variance for Naive CP, RAPS, and MC-CP on the five datasets. The results on CIFAR-10 show that Naive CP has the smallest mean set size, but this does not reflect its accuracy. Looking at the CIFAR-100 results, we can see that Naive CP has the smallest mean again, but its variance is substantially larger than the other results. In fact, we observed that Naive CP had set sizes ranging from $1$ to $86$, which indicates that the method cannot cope effectively with large-scale datasets with many (potential) classes. For both datasets, MC-CP achieves a smaller mean than RAPS and has less deviation around the mean. For CIFAR-100, RAPS has set sizes ranging from $33$ to $59$, whereas MC-CP has set sizes ranging from $30$ to $52$. These results show how the MC-CP can boost confidence in conformal prediction algorithms and achieve better results. Overall, we observe that advanced CP algorithms, like RAPS, tend to overestimate their predictions, and MC-CP reduces this overestimation.

\begin{figure}[t]
    \centerline{\includegraphics[width=0.95\linewidth]{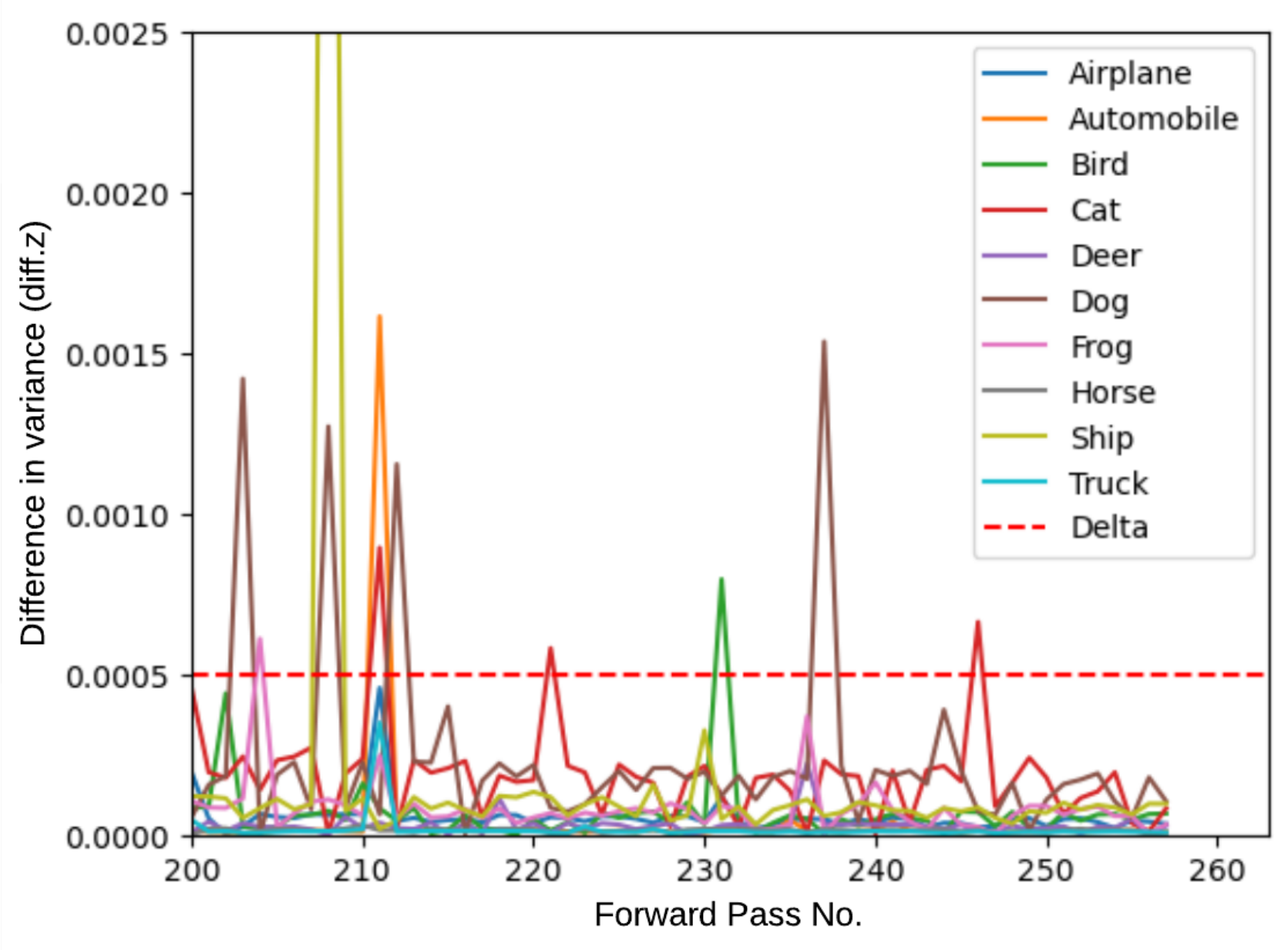}}
    \caption{Convergence of variance for each class during the Adaptive MC Dropout procedure.}
    \label{fig:variance-convergence}
\end{figure}

\begin{table}[t!]
\centering
\resizebox{\columnwidth}{!}{%
\begin{tabular}{|l|l|l|l|}
\hline
\rowcolor[HTML]{FFFFFF} 
\multicolumn{1}{|c|}{\cellcolor[HTML]{FFFFFF}\textbf{Model}} & \multicolumn{1}{c|}{\cellcolor[HTML]{FFFFFF}\textbf{Technique}} & \textbf{Test Error} & \textbf{Prediction Sizes} \\ \hline
\rowcolor[HTML]{F3F3F3} 
\cellcolor[HTML]{F3F3F3} & Naive & \cellcolor[HTML]{EFEFEF}5.16 $\pm$ 0.28 & \cellcolor[HTML]{EFEFEF}32.39 $\pm$ 23.44 \\ \cline{2-4} 
\rowcolor[HTML]{F3F3F3} 
\cellcolor[HTML]{F3F3F3} & RAPS & \cellcolor[HTML]{EFEFEF}4.90 $\pm$ 0.37 & \cellcolor[HTML]{EFEFEF}42.18 $\pm$ 5.17 \\ \cline{2-4} 
\rowcolor[HTML]{F3F3F3} 
\multirow{-3}{*}{\cellcolor[HTML]{F3F3F3}\textbf{VGG16}} & MC-CP & \cellcolor[HTML]{EFEFEF}3.86 $\pm$ 0.26 & \cellcolor[HTML]{EFEFEF}40.44 $\pm$ 5.09 \\ \hline
\rowcolor[HTML]{FFFFFF} 
\cellcolor[HTML]{FFFFFF} & Naive & \cellcolor[HTML]{FFFFFF}4.84 $\pm$ 0.34 & \cellcolor[HTML]{FFFFFF}28.40 $\pm$ 22.53 \\ \cline{2-4} 
\rowcolor[HTML]{FFFFFF} 
\cellcolor[HTML]{FFFFFF} & RAPS & \cellcolor[HTML]{FFFFFF}4.77 $\pm$ 0.19 & \cellcolor[HTML]{FFFFFF}36.50 $\pm$ 4.90 \\ \cline{2-4} 
\rowcolor[HTML]{FFFFFF} 
\multirow{-3}{*}{\cellcolor[HTML]{FFFFFF}\textbf{VGG19}} & MC-CP & \cellcolor[HTML]{FFFFFF}3.78 $\pm$ 0.14 & \cellcolor[HTML]{FFFFFF}32.48 $\pm$ 5.22 \\ \hline
\end{tabular}%
}
\caption{Test errors (\%) and prediction sizes per UQ method for two large DL models on the Tiny ImageNet dataset.}
\label{tab:large-model-results}
\end{table}

We also demonstrate that our MC-CP method works well with models at scale by assessing its capabilities using the VGG16 and VGG19 models on the Tiny ImageNet dataset. Table~\ref{tab:large-model-results} shows the reduced prediction set sizes for these models. The results on the larger DL models align with those shown in Table~\ref{tab:test-errors}, except in smaller magnitudes.

\vspace{1mm}\noindent
\textbf{Accuracy of Classes.} We next validated that MC-CP 
was not just doing significantly better than other methods in one or two classes but that indeed performs better for nearly all classes. Table~\ref{tab:classes-accuracy} shows the mean accuracy for all methods for each class in the CIFAR-10 dataset. We again see the trend where MC-CP increases the accuracy in comparison to the other methods, and the deviation between runs is also reduced. 
MC-CP consistently achieves an accuracy of approximately 97-99\%, showing that it does improve general accuracy, not just of a few classes. The \textit{Frog} class is the sole outlier where Naive CP achieves a higher accuracy, but this appears to be an outlier for that model; MC-CP still achieves a high mean accuracy of $99.02\% \pm 0.59$.

\begin{table*}[t]
\renewcommand{\arraystretch}{1.2}
\resizebox{\textwidth}{!}{%
\begin{tabular}{|l|llllllllll|}
\hline
 & \multicolumn{10}{c|}{\textbf{Accuracy of Class}} \\ \hline
\multicolumn{1}{|c|}{\textbf{Tech.}} & \multicolumn{1}{l|}{Airplane} & \multicolumn{1}{l|}{Automob.} & \multicolumn{1}{l|}{Bird} & \multicolumn{1}{l|}{Cat} & \multicolumn{1}{l|}{Deer} & \multicolumn{1}{l|}{Dog} & \multicolumn{1}{l|}{Frog} & \multicolumn{1}{l|}{Horse} & \multicolumn{1}{l|}{Ship} & Truck \\ \hline
\rowcolor[HTML]{F3F3F3} 
Baseline & \multicolumn{1}{l|}{\cellcolor[HTML]{F3F3F3}62.9$\pm$4.5} & \multicolumn{1}{l|}{\cellcolor[HTML]{F3F3F3}76.3 $\!\pm\!$6.9} & \multicolumn{1}{l|}{\cellcolor[HTML]{F3F3F3}39.8 $\!\pm\!$4.1} & \multicolumn{1}{l|}{\cellcolor[HTML]{F3F3F3}40.8 $\!\pm\!$2.9} & \multicolumn{1}{l|}{\cellcolor[HTML]{F3F3F3}59.8 $\!\pm\!$9.2} & \multicolumn{1}{l|}{\cellcolor[HTML]{F3F3F3}49.5 $\!\pm\!$6.2} & \multicolumn{1}{l|}{\cellcolor[HTML]{F3F3F3}87.8 $\!\pm\!$2.9} & \multicolumn{1}{l|}{\cellcolor[HTML]{F3F3F3}61.2 $\!\pm\!$5.3} & \multicolumn{1}{l|}{\cellcolor[HTML]{F3F3F3}76.8 $\!\pm\!$5.1} & 66.7 $\!\pm\!$3.4 \\ \hline
MC & \multicolumn{1}{l|}{{\color[HTML]{000000} 65.9 $\!\pm\!$5.3}} & \multicolumn{1}{l|}{75.4 $\!\pm\!$7.9} & \multicolumn{1}{l|}{45.3 $\!\pm\!$5.1} & \multicolumn{1}{l|}{39.4 $\!\pm\!$3.2} & \multicolumn{1}{l|}{57.3 $\!\pm\!$3.3} & \multicolumn{1}{l|}{59.2 $\!\pm\!$7.9} & \multicolumn{1}{l|}{76.4 $\!\pm\!$6.9} & \multicolumn{1}{l|}{72.1 $\!\pm\!$3.1} & \multicolumn{1}{l|}{77.1 $\!\pm\!$6.4} & 73.1 $\!\pm\!$6.9 \\ \hline
\rowcolor[HTML]{F3F3F3} 
Naive & \multicolumn{1}{l|}{\cellcolor[HTML]{F3F3F3}{\color[HTML]{000000} 93.5 $\!\pm\!$1.0}} & \multicolumn{1}{l|}{\cellcolor[HTML]{F3F3F3}{\color[HTML]{000000} 96.8 $\!\pm\!$0.7}} & \multicolumn{1}{l|}{\cellcolor[HTML]{F3F3F3}{\color[HTML]{000000} 92.4 $\!\pm\!$2.2}} & \multicolumn{1}{l|}{\cellcolor[HTML]{F3F3F3}{\color[HTML]{000000} 93.1 $\!\pm\!$2.2}} & \multicolumn{1}{l|}{\cellcolor[HTML]{F3F3F3}{\color[HTML]{000000} 97.6 $\!\pm\!$0.8}} & \multicolumn{1}{l|}{\cellcolor[HTML]{F3F3F3}{\color[HTML]{000000} 93.1 $\!\pm\!$0.8}} & \multicolumn{1}{l|}{\cellcolor[HTML]{F3F3F3}{\color[HTML]{000000} \textbf{99.3 $\!\pm\!$0.4}}} & \multicolumn{1}{l|}{\cellcolor[HTML]{F3F3F3}{\color[HTML]{000000} 91.4 $\!\pm\!$2.9}} & \multicolumn{1}{l|}{\cellcolor[HTML]{F3F3F3}{\color[HTML]{000000} 96.6 $\!\pm\!$0.8}} & {\color[HTML]{000000} 94.8 $\!\pm\!$1.4} \\ \hline
RAPS & \multicolumn{1}{l|}{{\color[HTML]{000000} 95.6 $\!\pm\!$1.2}} & \multicolumn{1}{l|}{{\color[HTML]{000000} 97.7 $\!\pm\!$0.7}} & \multicolumn{1}{l|}{{\color[HTML]{000000} 94.7 $\!\pm\!$1.2}} & \multicolumn{1}{l|}{{\color[HTML]{000000} 97.8 $\!\pm\!$0.5}} & \multicolumn{1}{l|}{{\color[HTML]{000000} 98.6 $\!\pm\!$0.6}} & \multicolumn{1}{l|}{{\color[HTML]{000000} 97.2 $\!\pm\!$0.9}} & \multicolumn{1}{l|}{{\color[HTML]{000000} 98.8 $\!\pm\!$0.4}} & \multicolumn{1}{l|}{{\color[HTML]{000000} 95.5 $\!\pm\!$0.9}} & \multicolumn{1}{l|}{{\color[HTML]{000000} 97.2 $\!\pm\!$1.0}} & {\color[HTML]{000000} 95.9 $\!\pm\!$1.7} \\ \hline
\rowcolor[HTML]{F3F3F3} 
MC-CP & \multicolumn{1}{l|}{\cellcolor[HTML]{F3F3F3}{\color[HTML]{000000} \textbf{97.8 $\!\pm\!$0.9}}} & \multicolumn{1}{l|}{\cellcolor[HTML]{F3F3F3}{\color[HTML]{000000} \textbf{99.3 $\!\pm\!$0.4}}} & \multicolumn{1}{l|}{\cellcolor[HTML]{F3F3F3}{\color[HTML]{000000} \textbf{98.2 $\!\pm\!$0.8}}} & \multicolumn{1}{l|}{\cellcolor[HTML]{F3F3F3}{\color[HTML]{000000} \textbf{98.3 $\!\pm\!$0.5}}} & \multicolumn{1}{l|}{\cellcolor[HTML]{F3F3F3}{\color[HTML]{000000} \textbf{99.0 $\!\pm\!$0.3}}} & \multicolumn{1}{l|}{\cellcolor[HTML]{F3F3F3}{\color[HTML]{000000} \textbf{98.2 $\!\pm\!$0.3}}} & \multicolumn{1}{l|}{\cellcolor[HTML]{F3F3F3}{\color[HTML]{000000} 99.0 $\!\pm\!$0.6}} & \multicolumn{1}{l|}{\cellcolor[HTML]{F3F3F3}{\color[HTML]{000000} \textbf{98.6 $\!\pm\!$0.4}}} & \multicolumn{1}{l|}{\cellcolor[HTML]{F3F3F3}{\color[HTML]{000000} \textbf{98.2 $\!\pm\!$0.5}}} & {\color[HTML]{000000} \textbf{98.5 $\!\pm\!$0.3}} \\ \hline
\end{tabular}%
}
\caption{Mean accuracy (\%) of classes for each method on the CIFAR-10 dataset.}
\label{tab:classes-accuracy}
\end{table*}

\begin{table*}[t]
\resizebox{\textwidth}{!}{%
\begin{tabular}{|ll|lllll|}
\hline
\multicolumn{2}{|l|}{} & \multicolumn{5}{c|}{\textbf{Delta $\delta$}} \\ \hline
\multicolumn{1}{|l|}{\textbf{Patience $P$}} & \textbf{Metrics} & \multicolumn{1}{r|}{0.1} & \multicolumn{1}{r|}{0.01} & \multicolumn{1}{r|}{0.001} & \multicolumn{1}{r|}{0.0001} & \multicolumn{1}{r|}{0.00001} \\ \hline
\rowcolor[HTML]{F3F3F3} 
\multicolumn{1}{|l|}{\cellcolor[HTML]{F3F3F3}} & Test Error & \multicolumn{1}{l|}{\cellcolor[HTML]{F3F3F3}6.25 $\pm$ 0.43} & \multicolumn{1}{l|}{\cellcolor[HTML]{F3F3F3}4.00 $\pm$ 0.50} & \multicolumn{1}{l|}{\cellcolor[HTML]{F3F3F3}3.60 $\pm$ 0.48} & \multicolumn{1}{l|}{\cellcolor[HTML]{F3F3F3}2.00 $\pm$ 0.50} & 2.00 $\pm$ 0.50 \\ \cline{2-7} 
\rowcolor[HTML]{F3F3F3} 
\multicolumn{1}{|l|}{\cellcolor[HTML]{F3F3F3}} & Mean fwd passes & \multicolumn{1}{l|}{\cellcolor[HTML]{F3F3F3}3.77 $\pm$ 0.82} & \multicolumn{1}{l|}{\cellcolor[HTML]{F3F3F3}9.22 $\pm$ 3.16} & \multicolumn{1}{l|}{\cellcolor[HTML]{F3F3F3}52.71 $\pm$ 18.59} & \multicolumn{1}{l|}{\cellcolor[HTML]{F3F3F3}387.71 $\pm$ 150.13} & 985.02 $\pm$ 75.78 \\ \cline{2-7} 
\rowcolor[HTML]{F3F3F3} 
\multicolumn{1}{|l|}{\multirow{-3}{*}{\cellcolor[HTML]{F3F3F3}1}} & Mean set size & \multicolumn{1}{l|}{\cellcolor[HTML]{F3F3F3}5.99 $\pm$ 1.32} & \multicolumn{1}{l|}{\cellcolor[HTML]{F3F3F3}5.35 $\pm$ 1.39} & \multicolumn{1}{l|}{\cellcolor[HTML]{F3F3F3}5.34 $\pm$ 1.21} & \multicolumn{1}{l|}{\cellcolor[HTML]{F3F3F3}5.01 $\pm$ 150.13} & 4.21 $\pm$ 1.33 \\ \hline
\multicolumn{1}{|l|}{\cellcolor[HTML]{FFFFFF}} & \cellcolor[HTML]{FFFFFF}Test Error & \multicolumn{1}{l|}{3.80 $\pm$ 0.40} & \multicolumn{1}{l|}{3.40 $\pm$ 0.49} & \multicolumn{1}{l|}{2.40 $\pm$ 0.49} & \multicolumn{1}{l|}{1.67 $\pm$ 0.47} & 1.33 $\pm$ 0.47 \\ \cline{2-7} 
\multicolumn{1}{|l|}{\cellcolor[HTML]{FFFFFF}} & \cellcolor[HTML]{FFFFFF}Mean fwd passes & \multicolumn{1}{l|}{13.10 $\pm$ 1.50} & \multicolumn{1}{l|}{31.45 $\pm$ 10.84} & \multicolumn{1}{l|}{142.36 $\pm$ 56.79} & \multicolumn{1}{l|}{812.34 $\pm$ 211.21} & 1000.00 $\pm$ 0.00 \\ \cline{2-7} 
\multicolumn{1}{|l|}{\multirow{-3}{*}{\cellcolor[HTML]{FFFFFF}10}} & \cellcolor[HTML]{FFFFFF}Mean set size & \multicolumn{1}{l|}{4.80 $\pm$ 1.44} & \multicolumn{1}{l|}{5.30 $\pm$ 1.21} & \multicolumn{1}{l|}{5.26 $\pm$ 1.34} & \multicolumn{1}{l|}{4.11 $\pm$ 1.81} & 4.06 $\pm$ 1.48 \\ \hline
\rowcolor[HTML]{F3F3F3} 
\multicolumn{1}{|l|}{\cellcolor[HTML]{F3F3F3}} & Test Error & \multicolumn{1}{l|}{\cellcolor[HTML]{F3F3F3}2.67 $\pm$ 0.47} & \multicolumn{1}{l|}{\cellcolor[HTML]{F3F3F3}2.50 $\pm$ 0.50} & \multicolumn{1}{l|}{\cellcolor[HTML]{F3F3F3}1.75 $\pm$ 0.43} & \multicolumn{1}{l|}{\cellcolor[HTML]{F3F3F3}1.38 $\pm$ 0.48} & 1.50 $\pm$ 0.48 \\ \cline{2-7} 
\rowcolor[HTML]{F3F3F3} 
\multicolumn{1}{|l|}{\cellcolor[HTML]{F3F3F3}} & Mean fwd passes & \multicolumn{1}{l|}{\cellcolor[HTML]{F3F3F3}103.09 $\pm$ 1.60} & \multicolumn{1}{l|}{\cellcolor[HTML]{F3F3F3}156.77 $\pm$ 35.56} & \multicolumn{1}{l|}{\cellcolor[HTML]{F3F3F3}672.57 $\pm$ 180.16} & \multicolumn{1}{l|}{\cellcolor[HTML]{F3F3F3}1000.00 $\pm$ 0.00} & 1000.00 $\pm$ 0.00 \\ \cline{2-7} 
\rowcolor[HTML]{F3F3F3} 
\multicolumn{1}{|l|}{\multirow{-3}{*}{\cellcolor[HTML]{F3F3F3}100}} & Mean set size & \multicolumn{1}{l|}{\cellcolor[HTML]{F3F3F3}5.01 $\pm$ 1.43} & \multicolumn{1}{l|}{\cellcolor[HTML]{F3F3F3}5.27 $\pm$ 1.20} & \multicolumn{1}{l|}{\cellcolor[HTML]{F3F3F3}4.75 $\pm$ 1.32} & \multicolumn{1}{l|}{\cellcolor[HTML]{F3F3F3}4.08 $\pm$ 1.33} & 4.19 $\pm$ 1.20 \\ \hline
\end{tabular}%
}
\caption{Sensitivity analysis on various threshold $\delta$ and patience $P$ combinations on the CIFAR-10 dataset ($K=1000$).}
\label{tab:various-delta}
\end{table*}

\vspace{1mm}\noindent
\textbf{Adaptive MC Dropout.} 
Figure~\ref{fig:variance-convergence} shows the convergence in each class variance for an example image from the CIFAR-10 dataset. We observe that at approximately 200 forward passes, the variance difference of all classes is below the $\delta$ threshold, and the patience counter starts increasing with every new iteration. However, at approximately 205 forward passes, the variance difference for classes \textit{Ship} and \textit{Automobile} spikes above the threshold; this is due to the stochastic nature of MC dropout. After 246 forward passes, all classes drop below the threshold, and the MC-CP procedure finishes early ten iterations later.

We also performed a sensitivity analysis of adaptive MC dropout to assess the impact of the threshold $\delta$ and patience $P$ on its performance. Table~\ref{tab:various-delta} shows the various combinations of $\delta$ and $P$ values used in these experiments. As $P$ increases and $\delta$ decreases (from top left to bottom right), we notice an increase in the mean number of forward passes yielding a corresponding reduction in test error (i.e., accuracy increase) and prediction set size. As expected, for $\delta \!=\! 0.00001$, $P \!=\! 100$ (bottom right) we obtain the traditional MC dropout, where the forward passes equals $K\!=\!1000$.

Finally, we demonstrate that adaptive MC dropout can save resources by comparing its execution overheads against traditional MC Dropout for $K\!=\!1000$, $\delta$=5e-4, $P$=10. Traditional MC dropout performed all 1000 forward passes on  CIFAR-10, and each image inference took an average of $35.52 \pm 0.42$ seconds. Adaptive MC Dropout averaged $500.21 \pm 196.37$ passes on all images and took an average of $17.99 \pm 7.09$ seconds. The ability of our method to diagnose convergence led to $\approx 50\%$ faster execution, meaning that the other $\approx 500$ forward passes were not needed. Considering memory consumption, as expected, both methods use the same memory ($\approx$1.07GB/$\approx$1.08GB for regular/adaptive MC Dropout) when training a full model plus inference on a dataset.

\subsection*{Regression Results}
\textbf{Regression Accuracy and Coverage.} 
In deep quantile regression, the mean absolute error (MAE) provides the magnitude of errors between the predicted quantiles and the true quantiles. 
Since MAE is less sensitive to outliers, we use it instead of (root) mean squared error.
We also compute the empirical coverage, which measures how often the predicted quantiles contain the true statistical quantile. 
Similarly to image classification, the objective is for the posterior prediction set to contain the true quantile.
Table~\ref{tab:regression-results} shows the MAE end empirical coverage for four different methods on the Boston Housing, Abalone, Blog Feedback, Concrete Strength and Protein datasets. 
We evaluated MC-CP against a baseline deep quantile regressor, the same deep quantile regressor with MC dropout, and conformalized quantile regression (CQR), the state-of-the-art CP regression method.

Looking at MAE, the traditional deep quantile regression model performs best across the five datasets. However, it also has a very low empirical coverage percentage across all five datasets. For example, in the Boston Housing dataset, the true data points are included in the predicted quantile only 22\% of the time. 
Similarly, although MC dropout increases the coverage by a considerable amount across all datasets, this method consistently leads to a worse MAE overall. 
In fact, we observe a tradeoff between these two methods. A low MAE comes with a low coverage, whereas a high coverage induces a high MAE.

\begin{figure}[htb]
    \centerline{\includegraphics[width=\linewidth]{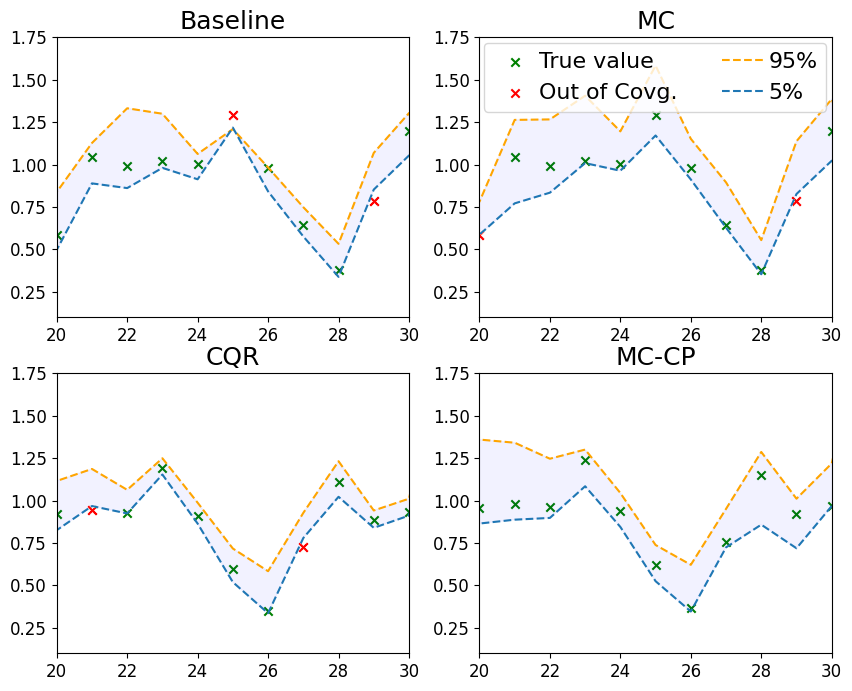}}
    \caption{Predicted quantiles (95\%, 5\%) of all four methods on a sample of the Boston Housing dataset.}
    \label{fig:quant-plots}
\end{figure}

Considering the CP-based methods, we observe that CQR provides the $1-\alpha$ coverage guarantee specified for all datasets, i.e., approximately $90\%$. Furthermore, CQR achieves this coverage with an MAE comparable to the baseline method in our experiments. Our MC-CP method reaches the highest empirical coverage across all four datasets, but it does this with slightly higher overall MAE (but lower standard deviation) on average than CQR. Given, however, the improved empirical coverage of MC-CP and its very close MAE results, we can conclude that MC-CP delivers very competitive results against the state-of-the-art CP method for regression. This is a particularly important insight, especially in safety-critical applications where higher coverage is vital. We conclude our evaluation with Figure~\ref{fig:quant-plots} which shows the predicted quantiles and coverage of the true values on an excerpt of the Boston Housing dataset. As expected, MC-CP yields slightly larger quantiles than CQR but has higher empirical coverage and misses fewer points.

\begin{table}[t]
\centering
\resizebox{\columnwidth}{!}{%
\begin{tabular}{|l|l|l|l|}
\hline
\multicolumn{1}{|c|}{\textbf{Dataset}} & \multicolumn{1}{c|}{\textbf{Technique}} & \multicolumn{1}{c|}{\textbf{MAE}} & \multicolumn{1}{c|}{\textbf{E. Coverage}} \\ \hline
\rowcolor[HTML]{F3F3F3} 
\cellcolor[HTML]{F3F3F3} & Baseline & 0.30 $\pm$ 0.02 & 23.52 $\pm$ 3.18 \\ \cline{2-4} 
\rowcolor[HTML]{F3F3F3} 
\cellcolor[HTML]{F3F3F3} & MC & 0.37 $\pm$ 0.02 & 72.83 $\pm$ 2.75 \\ \cline{2-4} 
\rowcolor[HTML]{F3F3F3} 
\cellcolor[HTML]{F3F3F3} & CQR & 0.31 $\pm$ 0.61 & 95.97$\pm$ 5.10 \\ \cline{2-4} 
\rowcolor[HTML]{F3F3F3} 
\multirow{-4}{*}{\cellcolor[HTML]{F3F3F3}\textbf{Boston Housing}} & {\color[HTML]{000000} MC-CP} & {\color[HTML]{000000} 0.35 $\pm$ 0.20} & {\color[HTML]{000000} 98.46 $\pm$ 4.83} \\ \hline
 & Baseline & 0.62 $\pm$ 0.04 & 47.86 $\pm$ 2.34 \\ \cline{2-4} 
 & MC & 0.64 $\pm$ 0.02 & 85.96 $\pm$ 1.82 \\ \cline{2-4} 
 & CQR & 0.62 $\pm$ 0.11 & 92.94 $\pm$ 2.36 \\ \cline{2-4} 
\multirow{-4}{*}{\textbf{Abalone}} & MC-CP & 0.64 $\pm$ 0.04 & 95.98 $\pm$ 3.07 \\ \hline
\rowcolor[HTML]{F3F3F3} 
\cellcolor[HTML]{F3F3F3} & Baseline & 2.12 $\pm$ 0.08 & 70.32 $\pm$ 5.70 \\ \cline{2-4} 
\rowcolor[HTML]{F3F3F3} 
\cellcolor[HTML]{F3F3F3} & MC & 2.61 $\pm$ 0.08 & 86.09 $\pm$ 5.80 \\ \cline{2-4} 
\rowcolor[HTML]{F3F3F3} 
\cellcolor[HTML]{F3F3F3} & CQR & 2.21 $\pm$ 0.10 & 90.73 $\pm$ 0.34 \\ \cline{2-4} 
\rowcolor[HTML]{F3F3F3} 
\multirow{-4}{*}{\cellcolor[HTML]{F3F3F3}\textbf{Blog Feedback}} & MC-CP & 2.40 $\pm$ 0.12 & 95.73 $\pm$ 0.34 \\ \hline
 & Baseline & 0.37 $\pm$ 0.01 & 20.55 $\pm$ 1.41 \\ \cline{2-4} 
 & MC & 0.54 $\pm$ 0.01 & 71.54 $\pm$ 5.51 \\ \cline{2-4} 
 & CQR & 0.37 $\pm$ 0.02 & 90.34 $\pm$ 3.69 \\ \cline{2-4} 
\multirow{-4}{*}{\textbf{Concrete}} & MC-CP & 0.44 $\pm$ 0.01 & 93.36 $\pm$ 2.49 \\ \hline
\rowcolor[HTML]{F3F3F3} 
\cellcolor[HTML]{F3F3F3} & Baseline & 1.35 $\pm$ 0.01 & 49.10 $\pm$ 1.75 \\ \cline{2-4} 
\rowcolor[HTML]{F3F3F3} 
\cellcolor[HTML]{F3F3F3} & MC & 1.49 $\pm$ 0.02 & 81.87 $\pm$ 0.21 \\ \cline{2-4} 
\rowcolor[HTML]{F3F3F3} 
\cellcolor[HTML]{F3F3F3} & CQR & 1.40 $\pm$ 0.02 & 94.79 $\pm$ 0.01 \\ \cline{2-4} 
\rowcolor[HTML]{F3F3F3} 
\multirow{-4}{*}{\cellcolor[HTML]{F3F3F3}\textbf{Protein}} & MC-CP & 1.45 $\pm$ 0.01 & 96.06 $\pm$ 0.73 \\ \hline
\end{tabular}%
}
\caption{Mean absolute error (MAE) and empirical coverage (\%) for each method on the Boston Housing, Abalone, Blog Feedback, Concrete Strength and Protein datasets.}
\label{tab:regression-results}
\end{table}

\begin{figure*}[hbt!]
    \centerline{\includegraphics[width=\linewidth]{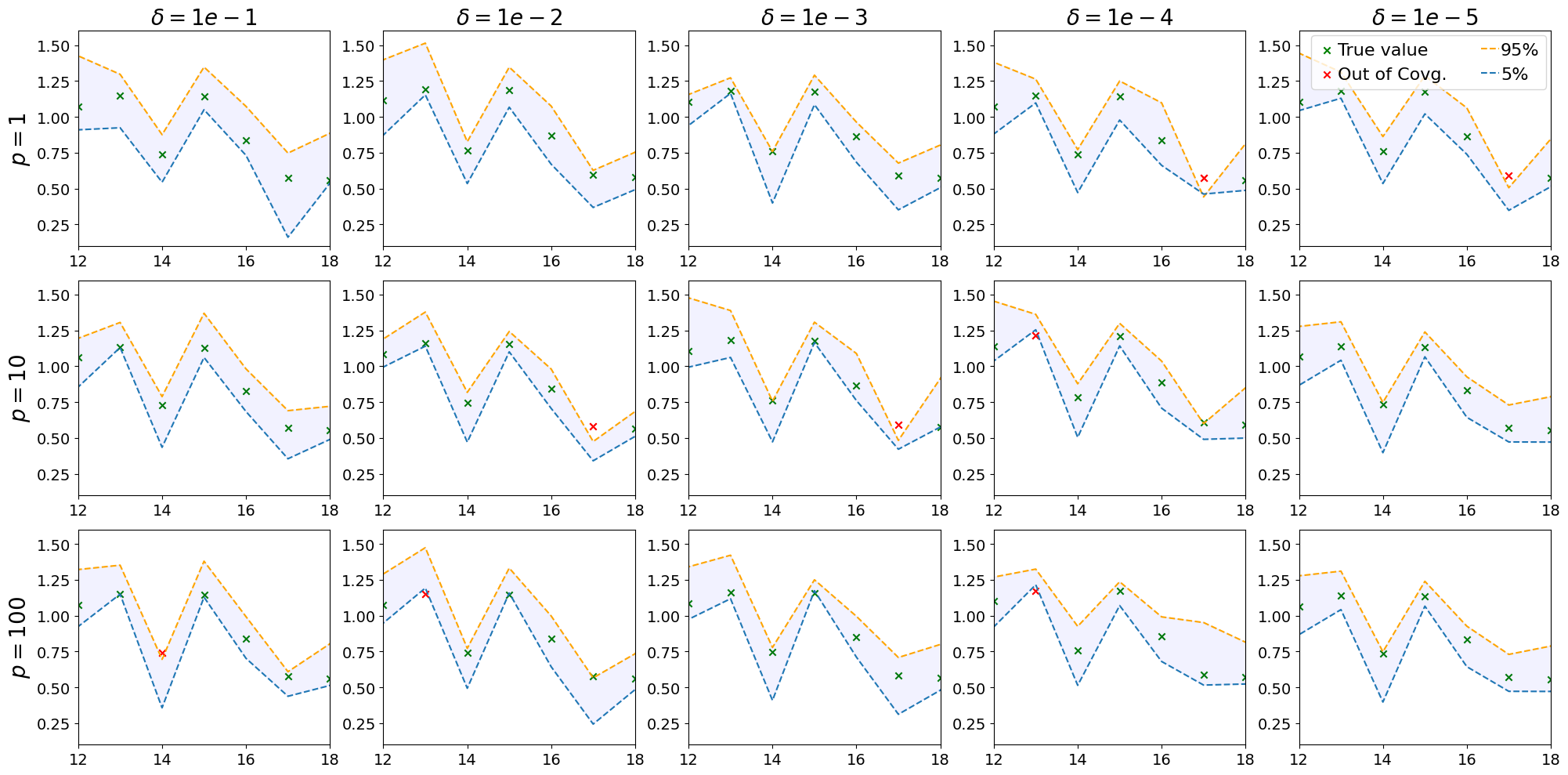}}
    \caption{Predicted quantiles (95\%, 5\%) for Table~\ref{tab:reg-sens-analysis} on a sample of the Boston Housing dataset.}
    \label{fig:delta-quant-table}
\end{figure*}

\vspace{1mm}\noindent
\textbf{Adaptive MC Dropout for Regression.} Similar to Table~\ref{tab:various-delta}, we performed sensitivity analysis on various combinations of $\delta$ and the patience value on deep quantile regression. Table~\ref{tab:reg-sens-analysis} shows how different combinations affect MAE and coverage. We also visualised the quantiles for the various combinations, which can be seen in Figure~\ref{fig:delta-quant-table}. Similarly to the results shown in Table~\ref{tab:various-delta}, a small $\delta$ and large patience show results comparable to traditional MC Dropout. It can be seen that with $\delta = 1e-5, p = 10$, we get considerable computational time saved with a compatible MAE to $\delta = 1e-5, p=100$.

\begin{table*}[htb]
\resizebox{\textwidth}{!}{%
\begin{tabular}{|ll|lllll|}
\hline
\multicolumn{2}{|l|}{} & \multicolumn{5}{c|}{\textbf{Delta}} \\ \hline
\multicolumn{1}{|l|}{\textbf{Patience}} & \textbf{Metrics} & \multicolumn{1}{r|}{1.00E-01} & \multicolumn{1}{r|}{1.00E-02} & \multicolumn{1}{r|}{1.00E-03} & \multicolumn{1}{r|}{1.00E-4} & \multicolumn{1}{r|}{1.00E-5} \\ \hline
\rowcolor[HTML]{F3F3F3} 
\multicolumn{1}{|l|}{\cellcolor[HTML]{F3F3F3}} & Mean MAE & \multicolumn{1}{l|}{\cellcolor[HTML]{F3F3F3}0.41 $\pm$ 0.02} & \multicolumn{1}{l|}{\cellcolor[HTML]{F3F3F3}0.40 $\pm$ 0.01} & \multicolumn{1}{l|}{\cellcolor[HTML]{F3F3F3}0.40 $\pm$ 0.02} & \multicolumn{1}{l|}{\cellcolor[HTML]{F3F3F3}0.40 $\pm$ 0.01} & 0.41 $\pm$ 0.03 \\ \cline{2-7} 
\rowcolor[HTML]{F3F3F3} 
\multicolumn{1}{|l|}{\cellcolor[HTML]{F3F3F3}} & Mean fwd passes & \multicolumn{1}{l|}{\cellcolor[HTML]{F3F3F3}3.00 $\pm$ 0.00} & \multicolumn{1}{l|}{\cellcolor[HTML]{F3F3F3}3.15 $\pm$ 0.84} & \multicolumn{1}{l|}{\cellcolor[HTML]{F3F3F3}4.83 $\pm$ 2.70} & \multicolumn{1}{l|}{\cellcolor[HTML]{F3F3F3}5.03 $\pm$ 3.03} & 5.14 $\pm$ 3.47 \\ \cline{2-7} 
\rowcolor[HTML]{F3F3F3} 
\multicolumn{1}{|l|}{\multirow{-3}{*}{\cellcolor[HTML]{F3F3F3}1}} & Mean Coverage & \multicolumn{1}{l|}{\cellcolor[HTML]{F3F3F3}84.85 $\pm$ 3.81} & \multicolumn{1}{l|}{\cellcolor[HTML]{F3F3F3}87.83 $\pm$ 2.17} & \multicolumn{1}{l|}{\cellcolor[HTML]{F3F3F3}88.85 $\pm$ 1.63} & \multicolumn{1}{l|}{\cellcolor[HTML]{F3F3F3}83.70 $\pm$ 9.78} & 87.50 $\pm$ 2.72 \\ \hline
\multicolumn{1}{|l|}{\cellcolor[HTML]{FFFFFF}} & \cellcolor[HTML]{FFFFFF}Mean MAE & \multicolumn{1}{l|}{0.39 $\pm$ 0.02} & \multicolumn{1}{l|}{0.38 $\pm$ 0.02} & \multicolumn{1}{l|}{0.37 $\pm$ 0.01} & \multicolumn{1}{l|}{0.36 $\pm$ 0.02} & 0.36 $\pm$ 0.01 \\ \cline{2-7} 
\multicolumn{1}{|l|}{\cellcolor[HTML]{FFFFFF}} & \cellcolor[HTML]{FFFFFF}Mean fwd passes & \multicolumn{1}{l|}{82.38 $\pm$ 209.43} & \multicolumn{1}{l|}{339.28 $\pm$ 368.87} & \multicolumn{1}{l|}{382.57 $\pm$ 393.45} & \multicolumn{1}{l|}{403.24 $\pm$ 399.50} & 459.85 $\pm$ 383.65 \\ \cline{2-7} 
\multicolumn{1}{|l|}{\multirow{-3}{*}{\cellcolor[HTML]{FFFFFF}10}} & \cellcolor[HTML]{FFFFFF}Mean Coverage & \multicolumn{1}{l|}{84.78 $\pm$ 14.13} & \multicolumn{1}{l|}{89.67 $\pm$ 0.54} & \multicolumn{1}{l|}{97.28 $\pm$ 0.54} & \multicolumn{1}{l|}{96.74 $\pm$ 2.17} & 97.28 $\pm$ 2.72 \\ \hline
\rowcolor[HTML]{F3F3F3} 
\multicolumn{1}{|l|}{\cellcolor[HTML]{F3F3F3}} & Mean MAE & \multicolumn{1}{l|}{\cellcolor[HTML]{F3F3F3}0.37 $\pm$ 0.01} & \multicolumn{1}{l|}{\cellcolor[HTML]{F3F3F3}0.34 $\pm$ 0.01} & \multicolumn{1}{l|}{\cellcolor[HTML]{F3F3F3}0.35 $\pm$ 0.01} & \multicolumn{1}{l|}{\cellcolor[HTML]{F3F3F3}0.35 $\pm$ 0.03} & 0.35 $\pm$ 0.01 \\ \cline{2-7} 
\rowcolor[HTML]{F3F3F3} 
\multicolumn{1}{|l|}{\cellcolor[HTML]{F3F3F3}} & Mean fwd passes & \multicolumn{1}{l|}{\cellcolor[HTML]{F3F3F3}484.33 $\pm$ 432.50} & \multicolumn{1}{l|}{\cellcolor[HTML]{F3F3F3}984.32 $\pm$ 98.02} & \multicolumn{1}{l|}{\cellcolor[HTML]{F3F3F3}974.85 $\pm$ 127.46} & \multicolumn{1}{l|}{\cellcolor[HTML]{F3F3F3}977.25 $\pm$ 113.95} & 982.10 $\pm$ 108.46 \\ \cline{2-7} 
\rowcolor[HTML]{F3F3F3} 
\multicolumn{1}{|l|}{\multirow{-3}{*}{\cellcolor[HTML]{F3F3F3}100}} & Mean Coverage & \multicolumn{1}{l|}{\cellcolor[HTML]{F3F3F3}97.83 $\pm$ 2.17} & \multicolumn{1}{l|}{\cellcolor[HTML]{F3F3F3}98.37 $\pm$ 1.63} & \multicolumn{1}{l|}{\cellcolor[HTML]{F3F3F3}93.76 $\pm$ 5.98} & \multicolumn{1}{l|}{\cellcolor[HTML]{F3F3F3}92.93 $\pm$ 5.98} & 95.65 $\pm$ 2.17 \\ \hline
\end{tabular}%
}
\caption{Sensitivity analysis on various hyperparameter combinations on the Boston Housing dataset.}
\label{tab:reg-sens-analysis}
\end{table*}

Similarly to the computational overheads investigation performed in image classification, we evaluated the overheads of traditional MC Dropout against Adaptive MC Dropout with the same parameters. Traditional MC Dropout performed all 1000 forward passes on the Boston Housing dataset, and each image inference took an average of $34.08 \pm 1.51$ seconds. Adaptive MC Dropout averaged $502.58 \pm 56.94$ forward passes on all images and took an average of $16.58 \pm 2.91$ seconds. 
Accordingly, we have obtained evidence that Adaptive MC Dropout was $\approx 50\%$ faster again.

\section*{Conclusion and Future Work}
\label{sec:conclusion}
Quantifying uncertainty in Deep Learning models is vital, especially when they are deployed in safety-critical applications. 
We introduced MC-CP, a hybrid uncertainty quantification method that combines a novel adaptive Monte Carlo dropout, informed by a coverage criterion to save resources during inference, with conformal prediction.
MC-CP delivers robust prediction sets/intervals by exploiting the statistical efficiency of MC dropout and the distribution-free coverage guarantees of conformal prediction. 
Our evaluation in classification and regression benchmarks showed that MC-CP offers significant improvements over advanced methods, like MC dropout, RAPS and CQR.
Our future work includes: 
(i) enhancing MC-CP to support object detection and segmentation tasks;
(ii) performing a more extensive evaluation using larger benchmarks and DL models; and
(iii) extending MC-CP to encode risk-related aspects in its analysis.

\section*{Acknowledgments}
\label{sec:acknowledgments}
This research has received funding from the Doctoral Centre for Safe, Ethical and Secure Computing (SEtS) at the University of York, UK, the European Union’s Horizon projects SESAME and SOPRANO (grant agreements No 101017258 and 101120990, respectively), and the EPSRC project `UKRI TAS Node in Resilience’ (EP/V026747/1), and the Assuring Autonomy International Programme. RC's work has also been funded by the Institute for Software Engineering and Software Technology `Jose Mar\'{i}a Troya Linero' at the University of M\'{a}laga.

\bibliography{aaai24}

\begin{thebibliography}{30}
\providecommand{\natexlab}[1]{#1}

\bibitem[{Abdar et~al.(2021)Abdar, Pourpanah, Hussain, Rezazadegan, Liu, Ghavamzadeh, Fieguth, Cao, Khosravi, Acharya et~al.}]{abdar2021review}
Abdar, M.; Pourpanah, F.; Hussain, S.; Rezazadegan, D.; Liu, L.; Ghavamzadeh, M.; Fieguth, P.; Cao, X.; Khosravi, A.; Acharya, U.~R.; et~al. 2021.
\newblock A review of uncertainty quantification in deep learning: Techniques, applications and challenges.
\newblock \emph{Information fusion}, 76: 243--297.

\bibitem[{Angelopoulos et~al.(2022)Angelopoulos, Bates, Malik, and Jordan}]{raps}
Angelopoulos, A.; Bates, S.; Malik, J.; and Jordan, M.~I. 2022.
\newblock Uncertainty Sets for Image Classifiers using Conformal Prediction.
\newblock arXiv:2009.14193.

\bibitem[{Angelopoulos and Bates(2022)}]{conformal-prediction-gentle-intro}
Angelopoulos, A.~N.; and Bates, S. 2022.
\newblock A Gentle Introduction to Conformal Prediction and Distribution-Free Uncertainty Quantification.
\newblock arXiv:2107.07511.

\bibitem[{Barnard et~al.(2003)Barnard, Duygulu, Forsyth, De~Freitas, Blei, and Jordan}]{image-anno-jmlr}
Barnard, K.; Duygulu, P.; Forsyth, D.; De~Freitas, N.; Blei, D.~M.; and Jordan, M.~I. 2003.
\newblock Matching words and pictures.
\newblock \emph{The Journal of Machine Learning Research}, 3: 1107--1135.

\bibitem[{Buza(2014)}]{blog-feedback}
Buza, K. 2014.
\newblock {BlogFeedback}.
\newblock UCI Machine Learning Repository.
\newblock {DOI}: https://doi.org/10.24432/C58S3F.

\bibitem[{Calinescu et~al.(2018)Calinescu, {\v{C}}e{\v{s}}ka, Gerasimou, Kwiatkowska, and Paoletti}]{calinescu2018efficient}
Calinescu, R.; {\v{C}}e{\v{s}}ka, M.; Gerasimou, S.; Kwiatkowska, M.; and Paoletti, N. 2018.
\newblock Efficient synthesis of robust models for stochastic systems.
\newblock \emph{Journal of Systems and Software}, 143: 140--158.

\bibitem[{de~Grancey et~al.(2022)de~Grancey, Adam, Alecu, Gerchinovitz, Mamalet, and Vigouroux}]{cp-obj-det}
de~Grancey, F.; Adam, J.-L.; Alecu, L.; Gerchinovitz, S.; Mamalet, F.; and Vigouroux, D. 2022.
\newblock Object Detection with Probabilistic Guarantees: A Conformal Prediction Approach.
\newblock In Trapp, M.; Schoitsch, E.; Guiochet, J.; and Bitsch, F., eds., \emph{Computer Safety, Reliability, and Security. SAFECOMP 2022 Workshops}, Lecture Notes in Computer Science, 316–329. Cham: Springer International Publishing.
\newblock ISBN 978-3-031-14862-0.

\bibitem[{Fan, Ge, and Mukherjee(2023)}]{cp-overestimate}
Fan, J.; Ge, J.; and Mukherjee, D. 2023.
\newblock UTOPIA: Universally Trainable Optimal Prediction Intervals Aggregation.
\newblock arXiv:2306.16549.

\bibitem[{Gal and Ghahramani(2016)}]{mcdropout}
Gal, Y.; and Ghahramani, Z. 2016.
\newblock Dropout as a bayesian approximation: Representing model uncertainty in deep learning.
\newblock In \emph{international conference on machine learning}, 1050–1059. PMLR.

\bibitem[{Gerasimou et~al.(2020)Gerasimou, Eniser, Sen, and Cakan}]{gerasimou2020importance}
Gerasimou, S.; Eniser, H.~F.; Sen, A.; and Cakan, A. 2020.
\newblock Importance-driven deep learning system testing.
\newblock In \emph{Proceedings of the ACM/IEEE 42nd International Conference on Software Engineering}, 702--713.

\bibitem[{Harrison and Rubinfield(1978)}]{boston-housing}
Harrison, D.; and Rubinfield, D.~L. 1978.
\newblock The Boston house-price data.
\newblock \url{http://lib.stat.cmu.edu/datasets/boston}.
\newblock Accessed: 2023-07-04.

\bibitem[{Kendall, Badrinarayanan, and Cipolla(2016)}]{bayesian-segnet}
Kendall, A.; Badrinarayanan, V.; and Cipolla, R. 2016.
\newblock Bayesian SegNet: Model Uncertainty in Deep Convolutional Encoder-Decoder Architectures for Scene Understanding.
\newblock ArXiv:1511.02680 [cs], arXiv:1511.02680.

\bibitem[{Kingma, Salimans, and Welling(2015)}]{gaussian-dropout}
Kingma, D.~P.; Salimans, T.; and Welling, M. 2015.
\newblock Variational Dropout and the Local Reparameterization Trick.
\newblock In Cortes, C.; Lawrence, N.; Lee, D.; Sugiyama, M.; and Garnett, R., eds., \emph{Advances in Neural Information Processing Systems}, volume~28. Curran Associates, Inc.

\bibitem[{Krizhevsky(2009)}]{CIFAR}
Krizhevsky, A. 2009.
\newblock \emph{Learning Multiple Layers of Features from Tiny Images}.
\newblock Ph.D. thesis, University of Tront.

\bibitem[{Kumar et~al.(2020)Kumar, Sahrawat, Maheshwari, Mahata, Stent, Yin, Shah, and Zimmermann}]{aaai-speech-recog}
Kumar, Y.; Sahrawat, D.; Maheshwari, S.; Mahata, D.; Stent, A.; Yin, Y.; Shah, R.~R.; and Zimmermann, R. 2020.
\newblock Harnessing gans for zero-shot learning of new classes in visual speech recognition.
\newblock In \emph{Proceedings of the AAAI Conference on Artificial Intelligence}, volume~34, 2645--2652.

\bibitem[{Lakshminarayanan, Pritzel, and Blundell(2017)}]{deep-ensembles}
Lakshminarayanan, B.; Pritzel, A.; and Blundell, C. 2017.
\newblock Simple and Scalable Predictive Uncertainty Estimation using Deep Ensembles.
\newblock In Guyon, I.; Luxburg, U.~V.; Bengio, S.; Wallach, H.; Fergus, R.; Vishwanathan, S.; and Garnett, R., eds., \emph{Advances in Neural Information Processing Systems}, volume~30. Curran Associates, Inc.

\bibitem[{LeCun et~al.(1998)LeCun, Bottou, Bengio, and Haffner}]{MNIST}
LeCun, Y.; Bottou, L.; Bengio, Y.; and Haffner, P. 1998.
\newblock Gradient-based learning applied to document recognition.
\newblock \emph{Proceedings of the IEEE}, 86(11): 2278--2324.

\bibitem[{MacKay(1992)}]{MacKay-BNN}
MacKay, D.~J. 1992.
\newblock A practical Bayesian framework for backpropagation networks.
\newblock \emph{Neural computation}, 4(3): 448–472.

\bibitem[{Missaoui, Gerasimou, and Matragkas(2023)}]{missaoui2023semantic}
Missaoui, S.; Gerasimou, S.; and Matragkas, N. 2023.
\newblock Semantic Data Augmentation for Deep Learning Testing using Generative AI.
\newblock In \emph{2023 38th IEEE/ACM International Conference on Automated Software Engineering (ASE)}, 1694--1698. IEEE.

\bibitem[{Moshkov et~al.(2020)Moshkov, Mathe, Kertesz-Farkas, Hollandi, and Horvath}]{tta-med2}
Moshkov, N.; Mathe, B.; Kertesz-Farkas, A.; Hollandi, R.; and Horvath, P. 2020.
\newblock Test-time augmentation for deep learning-based cell segmentation on microscopy images.
\newblock \emph{Scientific reports}, 10(1): 5068.

\bibitem[{Nash et~al.(1995)Nash, Sellers, Talbot, Cawthorn, and Ford}]{abalone}
Nash, W.; Sellers, T.; Talbot, S.; Cawthorn, A.; and Ford, W. 1995.
\newblock Abalone.
\newblock UCI Machine Learning Repository.
\newblock {DOI}: https://doi.org/10.24432/C55C7W.

\bibitem[{Pereira and Thomas(2020)}]{chall-ml}
Pereira, A.; and Thomas, C. 2020.
\newblock Challenges of machine learning applied to safety-critical cyber-physical systems.
\newblock \emph{Machine Learning and Knowledge Extraction}, 2(4): 579--602.

\bibitem[{Rana(2013)}]{protein}
Rana, P. 2013.
\newblock {Physicochemical Properties of Protein Tertiary Structure}.
\newblock UCI Machine Learning Repository.
\newblock {DOI}: https://doi.org/10.24432/C5QW3H.

\bibitem[{Romano, Patterson, and Candes(2019)}]{cqr}
Romano, Y.; Patterson, E.; and Candes, E. 2019.
\newblock Conformalized quantile regression.
\newblock \emph{Advances in neural information processing systems}, 32.

\bibitem[{Srivastava et~al.(2014)Srivastava, Hinton, Krizhevsky, Sutskever, and Salakhutdinov}]{dropout}
Srivastava, N.; Hinton, G.; Krizhevsky, A.; Sutskever, I.; and Salakhutdinov, R. 2014.
\newblock Dropout: A simple way to prevent neural networks from overfitting.
\newblock \emph{Journal of Machine Learning Research}, 15: 1929–1958.

\bibitem[{Vovk, Gammerman, and Shafer(2005)}]{cpbook}
Vovk, V.; Gammerman, A.; and Shafer, G. 2005.
\newblock \emph{Algorithmic learning in a random world}, volume~29.
\newblock Springer.

\bibitem[{Wang et~al.(2019)Wang, Li, Ourselin, and Vercauteren}]{tta-med}
Wang, G.; Li, W.; Ourselin, S.; and Vercauteren, T. 2019.
\newblock Automatic Brain Tumor Segmentation Using Convolutional Neural Networks with Test-Time Augmentation.
\newblock In Crimi, A.; Bakas, S.; Kuijf, H.; Keyvan, F.; Reyes, M.; and van Walsum, T., eds., \emph{Brainlesion: Glioma, Multiple Sclerosis, Stroke and Traumatic Brain Injuries}, 61--72. Cham: Springer International Publishing.
\newblock ISBN 978-3-030-11726-9.

\bibitem[{Wu, Zhang, and Xu(2017)}]{tinyImageNet}
Wu, J.; Zhang, Q.; and Xu, G. 2017.
\newblock Tiny imagenet challenge.
\newblock \emph{Technical report}.

\bibitem[{Xiao, Rasul, and Vollgraf(2017)}]{fashion-mnist}
Xiao, H.; Rasul, K.; and Vollgraf, R. 2017.
\newblock Fashion-MNIST: a Novel Image Dataset for Benchmarking Machine Learning Algorithms.
\newblock arXiv:1708.07747.

\bibitem[{Yeh(2007)}]{concrete}
Yeh, I.-C. 2007.
\newblock {Concrete Compressive Strength}.
\newblock UCI Machine Learning Repository.
\newblock {DOI}: https://doi.org/10.24432/C5PK67.

\end{thebibliography}

\end{document}